\documentclass[lettersize,journal]{IEEEtran}
\usepackage{amsmath,amsfonts}
\usepackage{algorithmic}
\usepackage{algorithm}
\usepackage{array}
\usepackage[caption=false,font=normalsize,labelfont=sf,textfont=sf]{subfig}
\usepackage{textcomp}
\usepackage{stfloats}
\usepackage{url}
\usepackage{verbatim}
\usepackage{graphicx}
\usepackage{cite}
\usepackage{listings}

\usepackage{microtype}
\usepackage{graphicx}
\usepackage{subcaption}
\usepackage{booktabs}
\usepackage{hyperref}
\lstdefinestyle{pythonappendix}{
    language=Python,
    basicstyle=\ttfamily\footnotesize, 
    keywordstyle=\color{blue},
    commentstyle=\color{green!60!black},
    stringstyle=\color{red},
    showstringspaces=false,
    frame=none,        
    numbers=left,              
    numberstyle=\tiny\color{gray}, 
    numbersep=5pt,
    breaklines=true,     
    captionpos=b,        
    aboveskip=5pt,       
    belowskip=5pt        
}
\usepackage[capitalize,noabbrev]{cleveref}
\newtheorem{theorem}{Theorem}[section]

\usepackage[textsize=tiny]{todonotes}

\begin{document}

\title{Interpretable and Sparse Linear Attention with Decoupled Membership-Subspace Modeling via MCR² Objective}

\author{Tianyuan Liu, Libin Hou, Linyuan Wang, Bin Yan}



\maketitle

\begin{abstract}
Maximal Coding Rate Reduction ($\mathrm{MCR}^2$)-driven white-box transformer, grounded in structured representation learning, unify interpretability and efficiency, providing a reliable white-box solution for visual modeling. However, in existing designs, tight coupling between “membership matrix $\boldsymbol{\Pi}$" and “subspace matrix $\boldsymbol{U}_{[K]}$" in $\mathrm{MCR}^2$ causes redundant coding under incorrect token projection. To this end, we decouple the functional relationship between the “membership matrix $\boldsymbol{\Pi}$" and “subspaces $\boldsymbol{U}_{[K]}$" in the $\mathrm{MCR}^2$ objective and derive an interpretable sparse linear attention operator from unrolled gradient descent of the optimized objective. 
Specifically, we propose to directly learn the membership matrix $\boldsymbol{\Pi}$ from inputs and subsequently derive sparse subspaces from the full space \(\boldsymbol{S}\).
Consequently, gradient unrolling of the optimized $\mathrm{MCR}^2$ objective yields an interpretable sparse linear attention operator: Decoupled Membership-Subspace Attention (DMSA). Experimental results on visual tasks show that simply replacing the attention module in Token Statistics Transformer(ToST) with DMSA(we refer to as DMST) not only achieves a faster coding reduction rate but also outperforms ToST by 1.08\%-1.45\% in top-1 accuracy on the ImageNet-1K dataset. Compared with vanilla Transformer architectures, DMST exhibits significantly higher computational efficiency and interpretability.
\end{abstract}

\section{Introduction}
\label{Introduction}
\IEEEPARstart{T}{ransformers} have cemented their status as the backbone of deep learning\cite{vaswani2017attention}—delivering state-of-the-art performance across speech recognition\cite{Dong2018Speech-Transformer}, computer vision\cite{liu2021swin}, natural language processing\cite{Devlin2018BERT,radford2018improving}, and other domains\cite{han2022survey,carion2020end,xie2021segformer,touvron2022deit}.
Despite their success, transformers suffer from poor interpretability; to address this key issue, the white-box Transformer architecture has recently emerged as a prominent research hotspot \cite{mcr2,chan2022redunet}.

The core idea of white-box Transformers lies in a mathematically driven design: unlike the empirical construction of traditional models, their network architectures are naturally generated by explicitly optimizing specific mathematical objective. The operation of each layer can be viewed as the incremental optimization step of the objective, enabling full mathematical interpretability. CRATE (Coding Rate Reduction Transformer) is a typical representative of white-box Transformers\cite{Yu2023CRATE}. With maximal Coding Rate Reduction ($\mathrm{MCR}^2$) as its core objectives, CRATE derives the self-attention layer as a gradient descent step for “compressing tokens" and the multi-layer perceptron (MLP) as a proximal operator for "sparsifying representations" by unrolling iterative optimization algorithms. This work has provided a new paradigm for the development of Transformer architectures.

Building on CRATE, the researchers further proposed the Token Statistics Transformer (ToST)\cite{ToST}. Inheriting the white-box architecture’s core advantages, ToST first derives a variational reformulation of the $\mathrm{MCR}^2$ objective and then unrolls the objective via gradient descent, giving rise to a new attention module called Token Statistics Self-Attention (TSSA). Unlike conventional attention mechanisms that rely on pairwise similarity calculations between tokens, this module captures global dependencies by modeling the second-moment statistical properties of tokens in low-dimensional subspaces. Inherently, it reduces the computational complexity to a linear scale (\(O(n)\)), thus achieving further improvements in both complexity efficiency and representational capability.

Notably, compared to the original CRATE model, ToST retains explicit modeling of the membership matrix $\boldsymbol{\Pi}$ in its objective, which quantifies the membership relationship between tokens and subspaces $\boldsymbol{U}_k$. However, in ToST, this membership relationship is strongly coupled with the subspace projection matrix $\boldsymbol{U}_k$: the computation of $\boldsymbol{\Pi}$ depends on the projection results of $\boldsymbol{U}_k$, while updates to $\boldsymbol{U}_k$ are constrained by the group assignments encoded in $\boldsymbol{\Pi}$. This one-way dependency easily leads to information misalignment, where the subspace structure learned by $\boldsymbol{U}_k$ mismatches the token assignments from $\boldsymbol{\Pi}$, ultimately introducing redundant coding and degrading the optimization precision of the $\mathrm{MCR}^2$ objective.

To address this core issue, we propose a decoupled optimization method for $\boldsymbol{\Pi}$ and $\boldsymbol{U}_k$. By refining the variational objective of $\mathrm{MCR}^2$, we break their strong coupling constraints, enabling independent optimization of subspaces and membership. Specifically, we directly learn the membership matrix $\boldsymbol{\Pi}$ via learnable weights; meanwhile, we introduce a concise yet effective soft thresholding operation to sparsify $\boldsymbol{\Pi}$, controlling the sparsity of subspaces and avoiding information misalignment caused by forced assignments between $\boldsymbol{U}_k$ and tokens. This yields DMSA(Decoupled Membership-Subspaces Attention) operator. By substituting the attention component in ToST with DMSA, we develop our DMST(Decoupled Membership-Subspaces Transformer) model. This decoupled design can both maintain effective learning of low-dimensional subspaces and guide each token to be accurately mapped to the target subspace during feature compression, ultimately achieving a dynamic balance between feature sparsity and representation accuracy.

\begin{figure}
\vskip 0.2in
    \centering
    \includegraphics[width=1\linewidth]{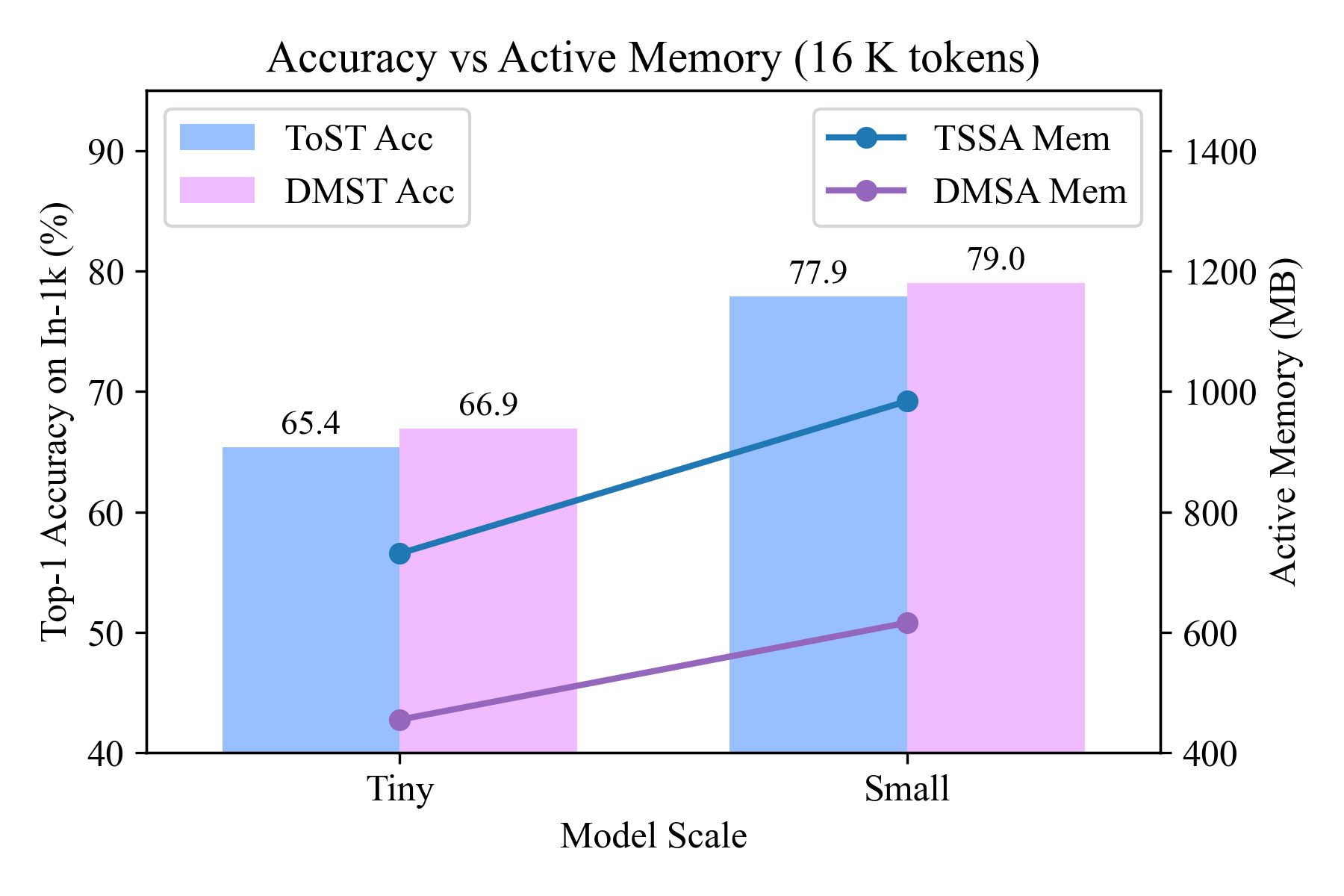}
    \caption{Top-1 accuracy on ImageNet-1K (left axis) and peak active memory under 16 K-token inputs (right axis).
Dark-blue solid line denotes TSSA memory; dark-purple solid line denotes DMSA memory.
Bars indicate validation accuracy across model scales. Our DMST model achieves better performance and consumes less memory.}
    \label{fig:acc}
\vskip -0.2in
\end{figure}

Moreover, experiments on visual tasks demonstrate that DMSA not only preserves linear complexity and white-box interpretability but also further enhances the ability of model representations, providing a more efficient and reliable solution for visual modeling tasks, as shown in Figure \ref{fig:acc}. 


We can summarize our contributions as below:
\begin{itemize}
\item We introduce DMSA, an inherently interpretable and efficient sparse linear attention operator, which is derived via gradient unrolling from an optimization objective that decouples membership $\boldsymbol{\Pi}$ from subspaces $\boldsymbol{U}_k$ within the $\mathrm{MCR}^2$.

\item We pioneer the decoupled design of membership matrix $\boldsymbol{\Pi}$ and subspaces $\boldsymbol{U}_k$, where $\boldsymbol{\Pi}$ is input-driven and sparse $\boldsymbol{U}_k$ is derived by activating the pre-learned full space $\boldsymbol{S}$ via $\boldsymbol{\Pi}$.

\item We prove that G1-type\footnote{A G1-type gating attention involves applying a gate after the head\cite{Qiu2025Gated},  which has been verified as the optimal gating position for attention mechanisms.} gating attention is a special variant of DMSA, which verifies that the essence of G1-type gating lies in adaptive optimal subspace $\boldsymbol{U}_k$ selection rather than mere weighting.

\item DMST achieves consistent performance gains across diverse visual benchmarks with negligible parameter overhead, while reducing peak memory usage by 21\%-validating its competitive efficacy.

\end{itemize}

\section{BACKGROUND AND PRELIMINARIES}
\label{BACKGROUND AND PRELIMINARIES}

\subsection{Notation}

In this paper, we adopt a set of notation commonly used in white-box Transformer design and information-theoretic feature processing. 

To obtain a useful representation of the input, we learn an encoder mapping $f:\mathbb{R}^{D \times N}\rightarrow  \mathbb{R}^{d \times n}$. The features that is, the output of the encoder are denoted by the variable $\boldsymbol{Z}=f(\boldsymbol{X})=\left[\boldsymbol{z_1}, \cdots, \boldsymbol{z_n}\right] \in \mathbb{R}^{d\times n}$, whence each $\boldsymbol{z_i} \in \mathbb{R}^d$ is a feature vector and specify these fed into the $l$-th attention layer as \( \boldsymbol{Z}^{l} \). Typically, the number of features \( n \) is the same as that of tokens \( N+1 \).

We use $R$ to present an estimate of the coding rate and $R^{\mathrm{c}}$ to present the coding rate under specific code. \(R_{c,f}\) is the upper bound with subspaces and $R_{\mathrm{cf}}^{\mathrm{var}}$ presents the variational form. $\boldsymbol{\Pi} = [\boldsymbol{\pi}_1, ..., \boldsymbol{\pi}_K] \in \mathbb{R}^{n \times K}$ denote a statistical “group assignment" matrix and \(\boldsymbol{U}_{[K]} = (\boldsymbol{U}_k)_{k \in [K]} \in (\mathbb{R}^{d \times p})^K\)presents learnable matrix in each attention layer.

\subsection{Representation Learning via Maximal Coding Rate Reduction}

Before presenting the optimization of our attention operator, we first recap the core ideas underlying the derivation of white-box attention. In practice, real-world data (e.g., images, videos, and text) are generally treated as realizations of a random variable $\boldsymbol{X}$, which is sampled from a high-dimensional probability distribution, yet such distributions inherently possess rich low-dimensional structural\cite{Fefferman2016Testing,Wright2022High-Dimensional}.
Previously, \cite{chan2022redunet} proposed to obtain parsimonious data representations by maximizing information gain\cite{mcr2}, a principled measure quantifying the information content of features. A concrete instantiation of this information gain is the coding rate reduction\cite{Yu2023CRATE}:

\begin{equation}
\Delta(R(\boldsymbol{Z}|\boldsymbol{\Pi}_{[K]}) = R(\boldsymbol{Z}) - R^{\mathrm{c}}(\boldsymbol{Z}|\boldsymbol{\Pi}_{[K]}).
\label{eq:mcr2}
\end{equation}

The first term $R(\boldsymbol{Z})$ is an estimate of the lossy coding rate for the whole set of features, when using a \textit{codebook} adapted to Gaussians. More specifically, if we view the token feature vectors $(\boldsymbol{z}_i)_{i \in [n]}$ in $\boldsymbol{Z} \in \mathbb{R}^{d \times n}$ as i.i.d. samples from a single zero-mean Gaussian, an approximation of their (lossy) coding rate, subject to quantization precision $\epsilon > 0$, is given in \cite{mcr2} as:

\begin{equation}
R(\boldsymbol{Z}) = \frac{1}{2} \log \det\!\left(\mathbf{I} + \alpha \boldsymbol{Z} \boldsymbol{Z}^*\right),\ \alpha = \frac{d}{n\epsilon^2}.
\end{equation}

The second term $R^{\mathrm{c}}$ in the rate reduction objective (\ref{eq:mcr2}) is also an estimate of the lossy coding rate, but under a different and more precise \textit{codebook}—one which views the token feature vectors $(\boldsymbol{z}_i)_{i \in [n]}$ as i.i.d. samples of a mixture of \textit{Gaussians}, where assignment of tokens to a particular Gaussian is known and specified by the Boolean membership matrices $\boldsymbol{\Pi}_{[K]} = (\boldsymbol{\pi}_k)_{k \in K}$. It assigns a membership probability to each token (i.e., the $i$-th row of $\boldsymbol{\Pi}$ corresponds to the membership probability vector of the $i$-th token). We thus obtain an estimate of the coding rate $R^{\mathrm{c}}$ as:

\begin{equation}
R^{\mathrm{c}}(\boldsymbol{Z} \mid \boldsymbol{\Pi}_{[K]}) = \frac{1}{2}\sum_{k=1}^K \log \det\!\left(\mathbf{I} + \gamma_k \boldsymbol{Z}\boldsymbol{\Pi}_k  \boldsymbol{Z}^*\right), \gamma_k = \frac{d}{n_k \epsilon^2}.
\end{equation}

\cite{Yu2023CRATE} further argues that the goal of representation learning is to learn a feature mapping (or representation) \(f \in \mathcal{F}: \mathbb{R}^{D \times N} \rightarrow \mathbb{R}^{d \times N}\), which decomposes the input data $\boldsymbol{X}$ into features $\boldsymbol{Z}$ with clear structure and compact compression. These features can be characterized by low-dimensional orthogonal subspace bases \(\boldsymbol{U}_{[K]}\), where the standard orthogonal basis \(\boldsymbol{U}_{[K]} = (\boldsymbol{U}_k)_{k \in [K]} \in (\mathbb{R}^{d \times p})^K\), and \(\boldsymbol{z}_i = \boldsymbol{U}_{s_i} \boldsymbol{\alpha}_i + \sigma \boldsymbol{w}_i\) for all \(i \in [n]\). Here, \(s_i\) denotes the index of the subspace that \(\boldsymbol{z}_i\) belongs to, and \(\boldsymbol{\alpha}_i\) is the encoding of \(\boldsymbol{z}_i\) corresponding to \(\boldsymbol{U}_{s_i}\).

\cite{Yu2023CRATE} then obtain an upper bound on the coding rate of token features $\boldsymbol{Z}$ by projecting tokens \(\boldsymbol{z}_i\) onto these subspaces and summing the coding rates across all subspaces. This upper bound is denoted as \(R_{c,f}\):

\begin{equation}
R_{c,f}(\boldsymbol{Z} \vert \boldsymbol{U}_{[K]}) = \frac{1}{2} \sum_{k=1}^K \log \det\left(\mathbf{I} + \beta (\boldsymbol{U}_k^\top \boldsymbol{Z})^* (\boldsymbol{U}_k^\top \boldsymbol{Z})\right)
\end{equation}

Subsequently, they propose the $\mathrm{MCR}^2$ objective to measure the learned representations: 

\begin{equation}
\max_{f \in \mathcal{F}} \Delta R_{c,f}(\boldsymbol{Z} \vert \boldsymbol{U}_{[K]}) = \max_{f \in \mathcal{F}} \left[ R(\boldsymbol{Z}) - R_{c,f}(\boldsymbol{Z} \vert \boldsymbol{U}_{[K]}) \right]
\end{equation}

Based on this objective framework, a fully mathematically interpretable Transformer-like deep network architecture (CRATE) can be constructed. In practice, CRATE-series models unroll the gradient of \(R_{c,f}(\boldsymbol{Z} \vert \boldsymbol{U}_{[K]})\) to derive a structural operator for layer-wise optimization: \(\nabla R_{c,f}(\boldsymbol{Z} \vert \boldsymbol{U}_{[K]}) = \text{MSSA}(\boldsymbol{Z} \vert \boldsymbol{U}_{[K]})\). This operator compresses the token set $\boldsymbol{Z}$, forcing it to distribute according to standard low-dimensional structures that match the inherent low-dimensionality of the data, ensuring the compactness of data coding.

\subsection{Building White-Box Token Statistics Self-Attention via Unrolled Optimization}
While CRATE matches the performance of the empirical Vision Transformer (ViT)\cite{Pixels2024An}, it exhibits quadratic time and memory complexity, which scales with the square of the token length $n$. \cite{ToST} performed a variational version of $\mathrm{MCR}^2$, thereby deriving a self-attention with linear complexity: Token Statistics Self-Attention (TSSA).

TSSA retain the explicit expression of the membership matrix in $R^{\mathrm{c}}(\boldsymbol{Z}, \boldsymbol{\Pi})$, and obtain a variational formulation via diagonalization:

\begin{equation}
\label{eq:RC(ZP)}
\begin{split}
R_{\mathrm{cf}}(\boldsymbol{Z}, \boldsymbol{\Pi})=\frac{1}{2}\sum_{k=1}^K \frac{n_k}{n}\log \det\!\left(\mathbf{I} + \frac{d}{\epsilon^2 n_k} \boldsymbol{Z}\mathrm{Diag}(\pi_k)\boldsymbol{Z}^\top\right),
\end{split}
\end{equation}


where $\boldsymbol{\pi}_k$ are the columns of $\boldsymbol{\Pi} = \boldsymbol{\Pi}(\boldsymbol{Z}|{\{\boldsymbol{U}_k\}^K_{k=1}})$ , which is defined below:  
\begin{equation}
\boldsymbol{\Pi} = \begin{bmatrix} \boldsymbol{\pi}_1^\top \\ \vdots \\ \boldsymbol{\pi}_K^\top \end{bmatrix} = \begin{bmatrix}
\mathrm{softmax}\!\left(\frac{1}{2\eta} \begin{bmatrix} \|\boldsymbol{U}_1^\top \boldsymbol{z}_1\|_2^2 \\ \vdots \\ \|\boldsymbol{U}_K^\top \boldsymbol{z}_1\|_2^2 \end{bmatrix}\right) \\
\vdots \\
\mathrm{softmax}\!\left(\frac{1}{2\eta} \begin{bmatrix} \|\boldsymbol{U}_1^\top \boldsymbol{z}_n\|_2^2 \\ \vdots \\ \|\boldsymbol{U}_K^\top \boldsymbol{z}_n\|_2^2 \end{bmatrix}\right)
\end{bmatrix}.
\end{equation}

%
Subsequently, the authors apply the "diagonal element inequality for orthogonal transformations of concave functions" to perform a variational reformulation of equation \ref{eq:RC(ZP)}, yielding the variational objective function as:

\begin{equation}
\begin{split}
&R_{\mathrm{cf}}^{\mathrm{var}}(\boldsymbol{Z}, \boldsymbol{\Pi} \mid \{\boldsymbol{U}_k\}_{k=1}^K) \\
&= \frac{1}{2}\sum_{k=1}^K \frac{n_k}{n}\sum_{i=1}^d f\left(\frac{1}{n_k}\left(\boldsymbol{U}_k^\top \boldsymbol{Z}\mathrm{Diag}(\pi_k)\boldsymbol{Z}^\top \boldsymbol{U}_k\right)_{ii}\right),
\end{split}
\end{equation}
where $f(x) = \log\!\left(1 + \frac{d}{\epsilon^2} x\right)$. After $\boldsymbol{U}_k$ is trained to satisfy the variational upper bound optimization requirement, it gains a concrete geometric meaning in the feature space: a learnable subspace matrix. 

In practice, the ToST model derived form $R_{\mathrm{cf}}^{\mathrm{var}}(\boldsymbol{Z}, \boldsymbol{\Pi} \mid \{\boldsymbol{U}_k\}_{k=1}^K)$ achieves superior performance over the original CRATE model. We argue that this gain originates from the explicit preservation of membership relations, which allows for a more complete coding representation. However, the ”membership matrix $\boldsymbol{\Pi}$” and ”subspace matrix $\boldsymbol{U}_{[K]}$” are subject to a tightly coupled joint learning regime; which may induce redundant coding when tokens are forced to project onto incorrect subspaces.
Guided by this insight, we aim to optimize the interaction between $\boldsymbol{\Pi}$ and $\boldsymbol{U}_{[K]}$ while retaining the essential learning of membership matrix.

\section{PROPOSED ATTENTION OPERATOR}
\label{sec:methology}
We now present the process of optimizing coding rates and refinement of the $\mathrm{MCR}^2$ objective, and propose a novel interpretable sparse linear attention operator: DMSA.


\subsection{An Optimization Form for Coding Rates and MCR$^2$}

To begin with, the membership matrix (also called the token-subspace grouping probability matrix) represents the probability that each token belongs to a specific subspace. TSSA establishes the relationship between $\boldsymbol{U}_k$ and $\boldsymbol{\Pi}$ via an $\ell_2$-norm functional relation, which essentially learns $\boldsymbol{U}_k$ to automatically derive $\boldsymbol{\Pi}$. However, when two variables have structural correlations but independent variation dimensions, this fixed relationship severely degrades expressiveness. In this case, decoupling their functional relation can eliminate redundant dependencies between variables, enabling the model to capture their intrinsic patterns more precisely, thereby improving optimization stability, generalization ability, and interpretability.

From the perspective of mutual information:
\begin{equation}
I(\boldsymbol{Z}; \boldsymbol{\pi}_k, \boldsymbol{U}_k) = I(\boldsymbol{Z}; \boldsymbol{\pi}_k) + I(\boldsymbol{Z}; \boldsymbol{U}_k \mid \boldsymbol{\pi}_k).
\end{equation}
When $I(\boldsymbol{Z}; \boldsymbol{\pi}_k)$ is maximized, additional information gain can be obtained by increasing $I(\boldsymbol{Z}; \boldsymbol{U}_k \mid \boldsymbol{\pi}_k)$. Thus, we propose optimizing the upper bound of the coding rate $R_{\mathrm{cf}}$ via the independent learning of the membership matrix $\boldsymbol{\Pi}$ independently.

Specifically, we characterize $R_{\mathrm{cf}}^{\mathrm{var}}(\boldsymbol{Z}, \boldsymbol{\pi}_k \mid \{\mathbf{U}_k\}_{k=1}^K)$ as:
\begin{equation}
\begin{split}
&R_{\mathrm{cf}}^{\mathrm{var}}(\boldsymbol{Z} \mid (\boldsymbol{\pi}_k, \boldsymbol{U}_k)) \\
&= \frac{1}{2}\sum_{k=1}^K \frac{n_k}{n}\sum_{i=1}^d f\left(\frac{1}{n_k}\left(\boldsymbol{U}_k^\top \boldsymbol{Z} \boldsymbol{\pi}_k \boldsymbol{Z}^\top \boldsymbol{U}_k\right)_{ii}\right).
\end{split}
\end{equation}

Following the variational derivation of \cite{ToST}, the gradient of $R_{\mathrm{cf}}^{\mathrm{var}}(\boldsymbol{Z} \mid (\boldsymbol{\pi}_k, \boldsymbol{U}_k))$ will be:
\begin{equation}
\begin{split}
&\nabla R_{\mathrm{cf}}^{\mathrm{var}}(\boldsymbol{Z} \mid (\boldsymbol{\pi}_k, \boldsymbol{U}_k)) \\
&= \frac{1}{n}\sum_{k=1}^K \boldsymbol{U}_k \mathrm{Diag}\left(\nabla f\left[(\boldsymbol{U}_k^\top \boldsymbol{Z})^{\odot2} \frac{\boldsymbol{\pi}_k }{\langle\boldsymbol{\pi}_k, \mathbf{1}\rangle}\right] \boldsymbol{U}_k^\top \boldsymbol{Z}\right) \mathrm{Diag}(\boldsymbol{\pi}_k).  
\end{split}
\end{equation}

Meanwhile, to establish the dynamic association between \(\boldsymbol{\pi}_k\) and the input \(\boldsymbol{Z}\), we set \(\boldsymbol{\pi}_k =  \boldsymbol{w}_k\boldsymbol{Z}\). Finally, we obtain:
\begin{equation}
\begin{split}
&\nabla R_{\mathrm{cf}}^{\mathrm{var}}(\boldsymbol{Z} \mid (\boldsymbol{\pi}_k, \boldsymbol{U}_k))\\
&= \frac{1}{n} \sum_{k=1}^K \boldsymbol{U}_k {D}(\boldsymbol{Z} \mid \boldsymbol{\pi}_k, \boldsymbol{U}_k) \boldsymbol{U}_k^\top \boldsymbol{Z} \mathrm{Diag}(\boldsymbol{\pi}_k),
\end{split}
\end{equation}

where

${D}(\boldsymbol{Z} \mid \boldsymbol{\pi}_k, \boldsymbol{U}_k) = \mathrm{Diag}\left( \nabla f\left[ (\boldsymbol{U}_k^\top \boldsymbol{Z})^{\odot 2} \frac{\boldsymbol{w}_k \boldsymbol{Z}}{\langle \boldsymbol{w}_k \boldsymbol{Z}, \mathbf{1} \rangle} \right] \right).$

After learning \(\boldsymbol{\pi}_k\), we can obtain a more flexible relationship between tokens and \(\boldsymbol{U}_k\). We observe that in the original setup, tokens and \(\boldsymbol{U}_k\) are assigned rigidly. \(\boldsymbol{\Pi} \in \mathbb{R}^{n \times K}\) computes the one-to-one correspondence between each token and the entire space. However, in real scenarios, some tokens do not belong to any learned subspace and some learned subspaces are not part of the entire space. Forcing such assignments introduces a "redundant coding rate". Moreover, this misaligned learning increases the redundancy of feature representations and reduces the discriminability of subspaces. When the metric relation between them is decoupled, \(\boldsymbol{\Pi}\) can learn more flexible assignment relations.

To reduce redundant coding, we introduce a unified low-rank manifold full space $\boldsymbol{S}$, where all subspaces \(\boldsymbol{U}_k\) are its sparse linear components. Mathematically, We regard $\boldsymbol{S}$ as a low-dimensional linear manifold spanned by a small number of latent bases in a high-dimensional space, and each \(\boldsymbol{U}_k\) is the $k$-th sparse substructure on $\boldsymbol{S}$, i.e., \(\boldsymbol{U}_k\) can be linearly represented by the basis vectors of $\boldsymbol{S}$ with sparse coefficients $\boldsymbol{\Pi}$. 

Here, the sparse selection of \(\boldsymbol{U}_k\) is determined by \(\boldsymbol{\Pi}\) through nonlinear activation: \(\boldsymbol{U}_k^S = \boldsymbol{S} \odot \sigma(\boldsymbol{\Pi})\) , where \(\sigma\) denotes a sparse activation function. Meanwhile, we adopt the soft thresholding operation ($\mathrm{ST}$) \cite{Su2024High-Similarity-Pass} as the activation function. The $\mathrm{ST}$ function can maintain the sum of non-zero components as 1 while controlling element sparsity:
\begin{equation}
\mathrm{ST}_j(s) = \max\left\{ s_j - \mathcal{K}(s), 0 \right\},
\end{equation}

where \(s_j\) is the $j$-th element of $s$, and \(\mathcal{K}(s): \mathbb{R}^N \to \mathbb{R}\) is a soft threshold function satisfying:
\begin{equation}
\sum_{j=1}^N \max\left\{ s_j - \mathcal{K}(s), 0 \right\} = 1.
\end{equation}
Substituting \(\boldsymbol{\Pi}\) and the $\mathrm{ST}$ function, we obtain:\(\boldsymbol{U}_k^S = \boldsymbol{S} \odot \mathrm{ST}(\boldsymbol{w}_k \boldsymbol{Z}).\) Finally, we derive the optimized upper bound of the coding rate as:
\begin{equation}
\begin{split}
&R_{\mathrm{cf}}^{\mathrm{var}}(\boldsymbol{Z} \mid (\boldsymbol{\pi}_k, \boldsymbol{U}_k^S))\\
&= \frac{1}{2} \sum_{k=1}^K \frac{n_k}{n} \sum_{i=1}^d f\left( \frac{1}{n_k} \left( (\boldsymbol{U}_k^S)^\top \boldsymbol{Z} \boldsymbol{\pi}_k \boldsymbol{Z}^\top \boldsymbol{U}_k^S \right)_{ii} \right).
\end{split}
\end{equation}


Subsequently, by substituting the optimized upper bound of the coding rate, we obtain:
\begin{equation}
\begin{split}
&\max_{f \in \mathcal{F}} \Delta R_{\mathrm{cf}}^{\mathrm{var}}(\boldsymbol{Z} \mid (\boldsymbol{\pi}_k, \boldsymbol{U}_k^S)) \\
&= \max_{f \in \mathcal{F}} \left[ R(\boldsymbol{Z}) - R_{\mathrm{cf}}^{\mathrm{var}}(\boldsymbol{Z} \mid (\boldsymbol{\pi}_k, \boldsymbol{U}_k^S)) \right]    
\end{split}
\end{equation}
We can theoretically prove that the decoupled membership-subspace objective yields a higher coding rate, as formalized in Theorem:

\begin{theorem}
  \label{thm:vari-mcr2}
  \begin{equation*}
  \begin{split}
  \Delta R_{\mathrm{cf}}^{\mathrm{var}}\bigl(\boldsymbol{Z} \mid (\boldsymbol{\pi}_k, \boldsymbol{U}_k^S)\bigr)
  &> \Delta R_{\mathrm{cf}}^{\mathrm{var}}\quad (\boldsymbol{Z}, \boldsymbol{\pi}_k \mid \boldsymbol{U}_k)
  \end{split}
  \end{equation*}
\end{theorem}

It provides a theoretical guarantee for our optimization. The detailed proof is provided in the appendix \ref{appendix:Theorem1}.

\subsection{Practical Implementation Considerations}
Following the $\mathrm{MCR}^2$ objective, we can perform feature compression via incremental iterative updates:
\begin{equation}
\begin{split}
    z_j^+ &= z_j - \nabla {R}_{\mathrm{cf}}^{\mathrm{var}}(\boldsymbol{Z} \mid (\boldsymbol{\pi}_k, \boldsymbol{U}_k^S))\\
    &= z_j - \frac{1}{n} \sum_{k=1}^K \boldsymbol{U}_k^S {D}(\boldsymbol{Z} \mid \boldsymbol{\pi}_k, \boldsymbol{U}_k^S) (\boldsymbol{U}_k^S)^\top \boldsymbol{Z} \mathrm{Diag}(\boldsymbol{\pi}_k). 
\end{split}
\end{equation}
We further convert the gradient operator \(\nabla {R}_{\mathrm{cf}}^{\mathrm{var}}(\boldsymbol{Z} \mid (\boldsymbol{\pi}_k, \boldsymbol{U}_k^S))\) into a network structure, yielding the attention operator that decouples membership relations and subspaces named the Decoupled Membership and Subspaces Attention (DMSA). The corresponding attention operator is defined as:
\begin{equation}
\begin{split}
&\mathrm{DMSA}\left( \boldsymbol{Z} \mid \{ \boldsymbol{\pi}_k, \boldsymbol{U}_k^S \}_{k=1}^K \right) \\
&= -\frac{1}{n} \sum_{k=1}^K \boldsymbol{U}_k^S {D}(\boldsymbol{Z} \mid \boldsymbol{\pi}_k, \boldsymbol{U}_k^S) (\boldsymbol{U}_k^S)^\top \boldsymbol{Z} \mathrm{Diag}(\boldsymbol{\pi}_k).    
\end{split}
\end{equation}
Specifically, we first train a full space $\boldsymbol{S}$, then a projection will be selected via \(\boldsymbol{\Pi}\): each attention head projects token features onto the basis of \(\boldsymbol{U}_k^S\) by multiplying with \(\boldsymbol{U}_k^S\), then multiplies with the diagonal matrix \({D}(\boldsymbol{Z} \mid \boldsymbol{\pi}_k, \boldsymbol{U}_k^S)\), projects back to the standard basis via \(\boldsymbol{U}_k^S\), and finally multiplies with \(\mathrm{Diag}(\boldsymbol{\pi}_k)\) to obtain the assignment value of each token. The core of our attention layer is decoupling the complex functional relation between \(\boldsymbol{\Pi}\) and \(\boldsymbol{U}_k\), while introducing the concept of a full projection space, it enables the model to flexibly select subspaces and implement soft assignment of membership relations. Additionally, the ordering of different elements in the same feature set (e.g., spatial layout of image features) directly affects the logical consistency of membership assignments. We adopt the idea of RoPE\cite{Su2024RoFormer} to apply positional encoding to the input features of \(\boldsymbol{\Pi}\), ensuring the consistency of membership assignments. Overall, we construct an alternative attention mechanism: by replacing the TSSA attention operator in ToST with the DMSA operator, we finally obtain the  Decoupled Membership and Subspaces Transformer(DMST). We present one layer of DMST in Figure \ref{fig:DMSA}, and provide the model pseudocode in Appendix \ref{appendix:pseudocode}.

\begin{figure*}[ht]
\vskip 0.2in
\begin{center}
    \centering
    \includegraphics[width=0.9\linewidth]{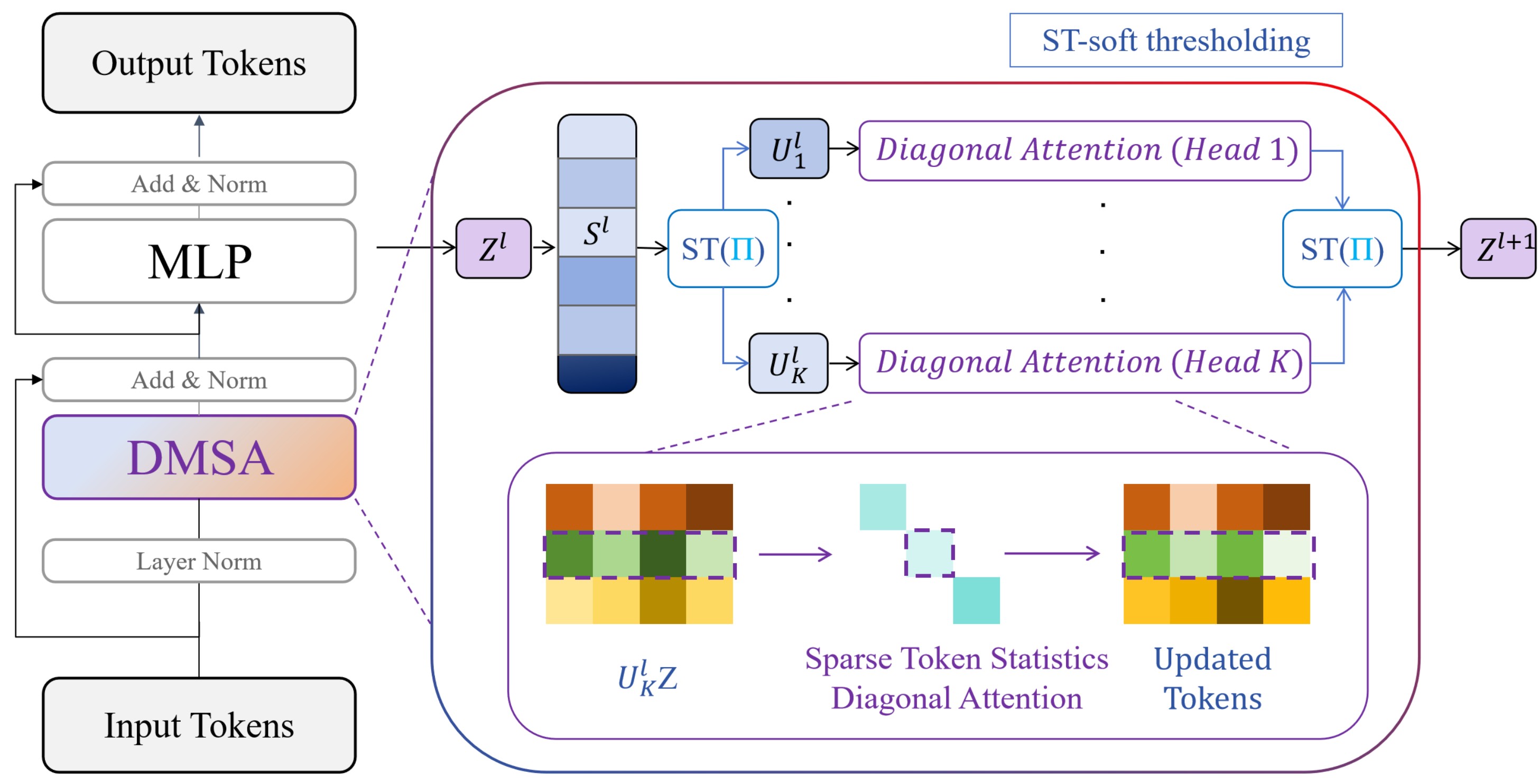}
    \caption{\textbf{One layer ${l}$ of the proposed Decoupled Membership and Subspaces Transformer(DMST).} Notably, the self-attention of DMST transforms tokens $\boldsymbol{Z}^l$ efficiently to $\boldsymbol{Z}^{l+1}$, via multiplying each row of the projected token by a scalar. The projected subspace is selected via an independent sparse activate membership $\boldsymbol{\Pi}$.}
    \label{fig:DMSA}
\end{center}
\vskip -0.2in
\end{figure*}

\subsection{DMSA Variants are Gate-controlled Attention}

In this section, we prove that DMSA can be transformed to be equivalent to the state-of-the-art gated attention method in \cite{Qiu2025Gated}. Specifically, if we remove the weighted computation of the membership matrix \(\boldsymbol{\Pi}\) and only retain the sparsification of \(\boldsymbol{U}_k^S\), DMSA reduces to channel attention\cite{CBSA}, which differs from standard attention in its selection of structural features, and uses the second-order moment of token projections for adaptive scaling: 
\begin{equation}
\begin{split}
&\sum_{k=1}^K \boldsymbol{U}_k^S \mathrm{D}_k (\boldsymbol{U}_k^S)^\top \boldsymbol{Z}, \\
&\quad \mathrm{D}_k = \mathrm{Diag}\left( \nabla {f}\left[ \boldsymbol{U}_k^\top \boldsymbol{Z}^{\odot 2} \frac{\boldsymbol{w}_k \boldsymbol{Z}}{(\boldsymbol{w}_k \boldsymbol{Z}, \mathbf{1})} \right] \right)     
\end{split}
\end{equation}

In this case, each selected subspace always corresponds to the principal direction of any input token. By substituting \(\boldsymbol{U}_k^S = \boldsymbol{S} \odot \sigma(\boldsymbol{\Pi}) = \boldsymbol{W}_S \odot \sigma(\boldsymbol{w}_k \boldsymbol{Z})\) , where \(\mathrm{G}_k = \sigma(\boldsymbol{\pi}_k) = \sigma(\boldsymbol{w}_k \boldsymbol{Z})\)), we derive:
\begin{equation}
\begin{split}
\sum_{k=1}^K \boldsymbol{U}_k^S \mathrm{D}_k (\boldsymbol{U}_k^S)^\top \boldsymbol{Z} &= \sum_{k=1}^K \left( \boldsymbol{S} \odot \sigma(\boldsymbol{\Pi}) \right) \mathrm{D}_k \left( \boldsymbol{S} \odot \sigma(\boldsymbol{\Pi}) \right)^\top \boldsymbol{Z} \\
&= \sum_{k=1}^K \left(\boldsymbol{W}_S \odot \mathrm{G}_k \right) \mathrm{D}_k \left( \boldsymbol{W}_S \odot \mathrm{G}_k \right)^\top \boldsymbol{Z}
\end{split}
\end{equation}

Converting the Hadamard product to matrix multiplication, we further simplify:
\begin{equation}
\begin{split}
&= \sum_{k=1}^K \mathrm{G}_k \left( w_S \mathrm{D}_k w_S^\top \boldsymbol{Z} \right) \\
&= \text{Non-Linearity-Map}\left( \sum_{k=1}^K w_S \mathrm{D}_k w_S^\top \boldsymbol{Z} \right)
\end{split}
\end{equation}
Here, \(\mathrm{G}_k\) denotes element-wise gating scaling. In this case, the sparse membership matrix acts as an additional gate in standard channel attention, which enhances the non-linearity of the output. This is the natural form of learning sparse mappings from manifolds, which also corroborates that the core mechanism underlying the effectiveness of G1-type gating lies in reducing redundant coding for spatial matching.

\begin{figure}[ht]
\vskip 0.2in
    \centering
    \includegraphics[width=0.3\linewidth]{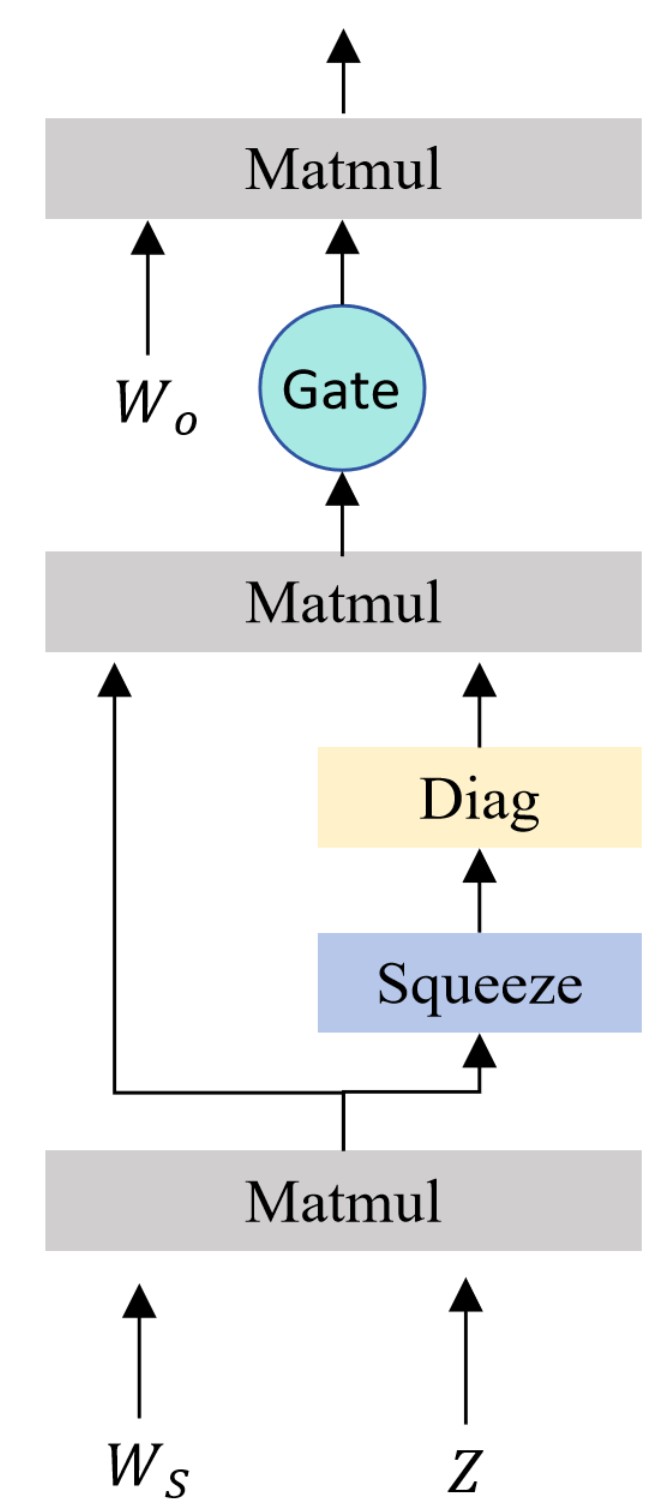}
    \caption{DMSA variants. The structure is theoretically equivalent to a form of gated channel attention.}
    \label{fig:dmsa-channel}
\vskip -0.2in
\end{figure}

\section{EXPERIMENTS}
In this section, we empirically evaluate the DMST's performance on real-world datasets and downstream tasks. As detailed in Section \ref{sec:methology}, DMSA is designed under the \(\mathrm{MCR}^2\) principle, with the compression objective instantiated directly from optimized mathematical formulations. We conduct experiments to demostrate that the DMST attention still optimize the intended compression objective in practice. Meanwhile, we perform experiments on ImageNet-1K classification\cite{krizhevsky2012imagenet}, CIFAR\cite{Krizhevsky2009Cifar}, Oxford-pets\cite{Oxford-pets}, Food101\cite{Food-101} and SVHN\cite{SVHN}. Additionally we confirm the contribution of individual DMST components through systematic ablation studies.

\subsection{Dataset and Optimization}

For visual classification tasks, we adopt the ImageNet-1K(In-1k), which comprises 1,000 object categories, 1.2 million training samples, and 50,000 validation samples. We first pre-trained our DMST model on ImageNet-1K; subsequently, we performed transfer learning experiments on a suite of small-scale natural image datasets, namely CIFAR10/100, Oxford-IIIT-Pets, Food-101, and SVHN. We report the top-1 validation accuracy of our model and compare it with the performance of the ToST baseline. Details on training and optimization are provided in Appendix \ref{appendix:training sets}.

\subsection{Result}

Table \ref{tab:dmst_accuracy} reports the top-1 classification accuracy of DMST models with varying parameter sizes: pre-trained on the In-1K, and fine-tuned on small-scale downstream datasets (CIFAR-10, CIFAR-100, Oxford-Pets, Food-101, and SVHN) for transfer learning. We define two scaled DMST models: DMST-T for the tiny-scale configuration and DMST-S for the small-scale configuration. The results indicate that DMST outperforms the ToST baseline significantly under identical parameter budgets, confirming the effectiveness of decoupling membership relations and subspace learning in the attention operator.

\begin{table}[t]
\caption{Top 1 accuracy of DMSA on various datasets with different model sizes}
\label{tab:dmst_accuracy}
\vskip 0.15in
\begin{center}
\begin{small}
\begin{sc}
\begin{tabular}{lcccc}
\toprule
Datasets & TSSA & DMST-T & TSSA & DMST-S \\
\cmidrule{1-5}
\#parameters  & 5.57M & 5.58M & 22.20M & 22.25M \\
\midrule
ImageNet  & 65.42\% & 66.87\% & 77.90\% & \textbf{78.98\%} \\
\midrule
CIFAR-10  & 93.30\% & 94.04\% & 95.19\% & \textbf{95.86\%} \\
CIFAR-100  & 76.37\% & 77.23\% & 82.20\% & \textbf{83.53\%} \\
Oxford-pets  & 81.57\% & 82.44\% & 81.87\% & \textbf{83.40\%} \\
Food101  & 81.97\% & 82.64\% & 83.40\% & \textbf{83.82\%} \\
SVHN  & 89.20\% & 90.68\% & 91.13\% & \textbf{91.26\%} \\
\bottomrule
\end{tabular}
\end{sc}
\end{small}
\end{center}
\vskip -0.1in
\end{table}

\begin{figure}[ht]
\vskip 0.2in
    \centering
    \includegraphics[width=1\linewidth]{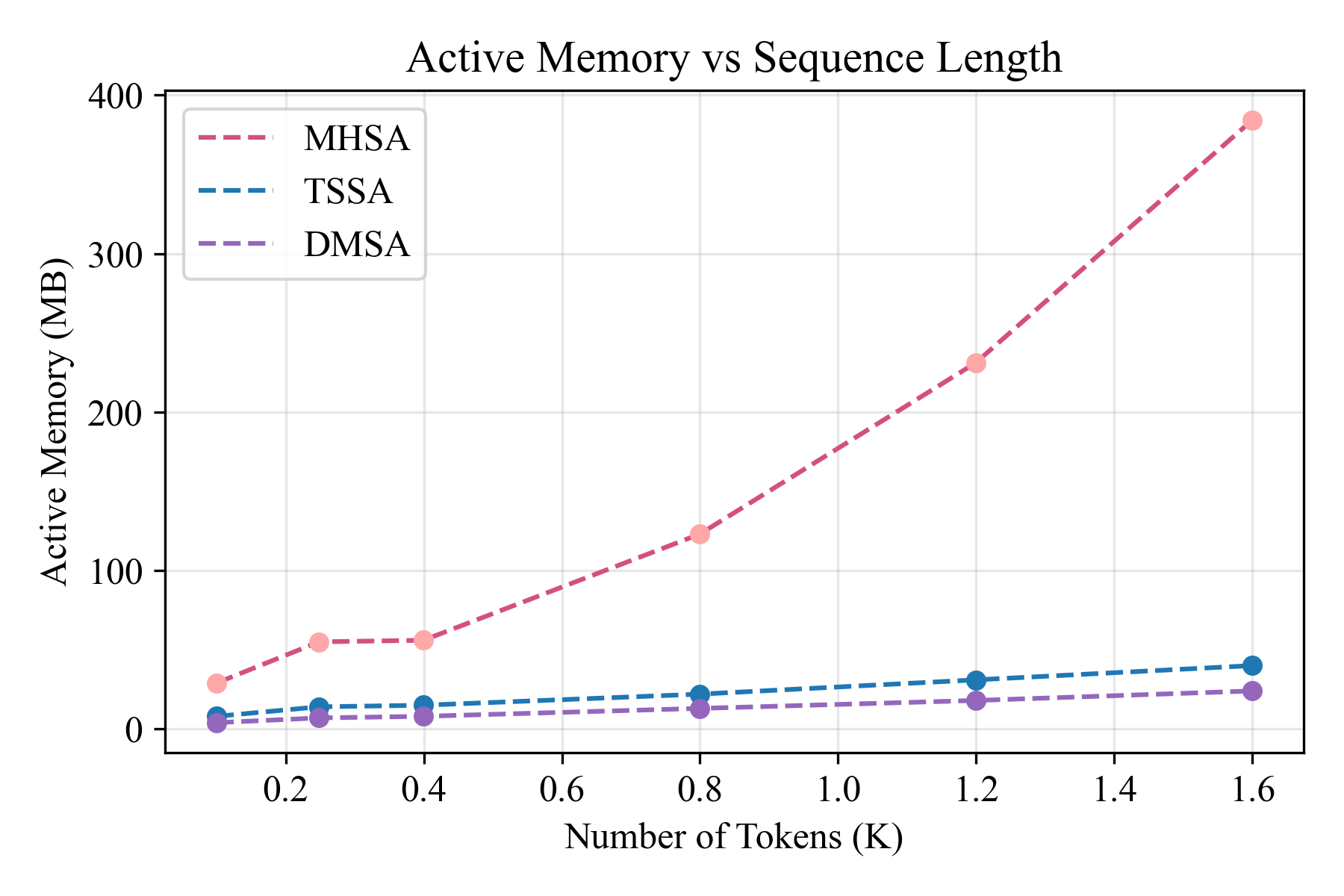}
    \caption{DMSA achieves lower memory usage than MHSA in ViT and linear attention TSSA.}
    \label{fig:compute memory}
\vskip 0.2in
\end{figure}

As illustrated in Figure~\ref{fig:compute memory}, we compare the activation memory consumption of MHSA(adopted in vanilla ViT), TSSA, and DMSA across different input sequence lengths. Our results reveal that MHSA exhibits a substantial increase in activation memory as the token count grows, the TSSA operator shows a linear growth trend in activation memory, while our proposed DMSA maintains the minimal activation memory throughout all tested lengths.

\subsection{Alblation study}
From the perspective of sparsity implementation, there are two essential approaches: one preserves the learning of \(\boldsymbol{U}_k\) and sparsifies tokens; the other (adopted in this work) sparsifies the full space: \(\boldsymbol{U}_k^S = \boldsymbol{S} \odot \sigma(\boldsymbol{\Pi})\). To investigate the differences between these two approaches, we conduct comparative experiments on different sparsity positions: TPUSA: Sparsify \(\boldsymbol{\Pi}\) along the token dimension and use it directly for computation; DMSA: Sparsify \(\boldsymbol{\Pi}\) along the subspace dimension, then sparsify the full space; TPSUSA: Sparsify \(\boldsymbol{\Pi}\) along both dimensions and involve it in computation. Table \ref{tab:sparsity_ablation} presents the ablation results of these models on CIFAR-10 and SVHN datasets. The results show that DMSA achieves the best performance.
\begin{table}[t]
\caption{Top 1 Accuracy of Models with Different Sparsity Dimensions}
\label{tab:sparsity_ablation}
\vskip 0.15in
\centering
\small
\begin{tabular}{lcccc}
\toprule
Datasets & TSSA & TPUSA  & DMSA  & TPSUSA \\
\cmidrule{1-5}
\#parameters & 3.19M & 3.20M & 3.20M & 3.20M \\
\midrule
CIFAR-10 & 61.59\% & 62.40\% & 62.62\% & 63.13\% \\
SVHN & 85.55\% & 86.25\% & 87.42\% & 87.09\% \\
\bottomrule
\end{tabular}
\vskip -0.1in
\end{table}

Analysis: Sparsifying \(\boldsymbol{\Pi}\) along the token ($n$) dimension causes some tokens to be excluded from compression during assignment, reducing the effective utilization rate. In contrast, sparsifying \(\boldsymbol{\Pi}\) along the subspace ($h$) dimension preserves token assignment, only poorly learned subspaces are discarded. This is an efficient strategy from the perspective of sparsity optimization.

We also compare the \(\sigma(\boldsymbol{\Pi})\) activation ($\mathrm{ST}$ sparsity activation) with common activation functions (Sigmoid, ReLU, GELU). The results (on CIFAR-10) show that $\mathrm{ST}$ activation for sparsifying \(\boldsymbol{U}_k\) yields the best performance.

\begin{table}[t]
\caption{Top 1 Accuracy of DMSA on CIFAR-10 with Different Activation Functions}
\label{tab:activation_ablation}
\vskip 0.15in
\centering
\small
\begin{tabular}{lccccc}
\toprule
Model & Sigmoid & ReLU & GELU & ST & ACC \\
\midrule
DMSA & $\checkmark$ & & & & 60.92\% \\
DMSA & & $\checkmark$ & & & 61.27\% \\
DMSA & & & $\checkmark$ & & 61.11\% \\
DMSA & & & & $\checkmark$ & 61.30\% \\
\bottomrule
\end{tabular}
\vskip -0.1in
\end{table}

\subsection{Layer-Wise Analysis of the DMSA}
\begin{figure}[ht]
\vskip -0.2in
    \centering
    \includegraphics[width=1\linewidth]{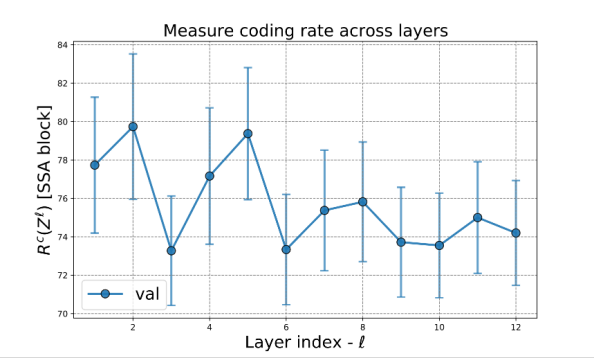}
    \caption{The variational compression term \({R}_{\mathrm{cf}}^{\mathrm{var}}(\boldsymbol{Z} \mid (\boldsymbol{\pi}_k, \boldsymbol{U}_k^S))\) of the DMSA outputs $\boldsymbol{Z}$.}
    \label{fig:layerwise}
\vskip 0.2in
\end{figure}

In this subsection, we demonstrate that DMSA positively contributes to the compression term, and the decoupled \(\boldsymbol{\Pi}\) matrix can maintain effective feature grouping. To verify the compression effect of DMSA, we measure the term \({R}_{\mathrm{cf}}^{\mathrm{var}}(\boldsymbol{Z} \mid (\boldsymbol{\pi}_k, \boldsymbol{U}_k^S))\) (for the output \(\boldsymbol{Z}\) after the DMSA block) at each layer of the DMST model. Specifically, we use 1,000 data samples sampled from the ImageNet-1K validation set: this term is evaluated at the sample level, and the average value is taken. Figure \ref{fig:layerwise} presents the optimization curve for the \(\mathrm{MCR}^2\) objective.

\begin{figure}[ht]
\vskip 0.2in
\begin{center}
    \centering{\includegraphics[width=1\linewidth]{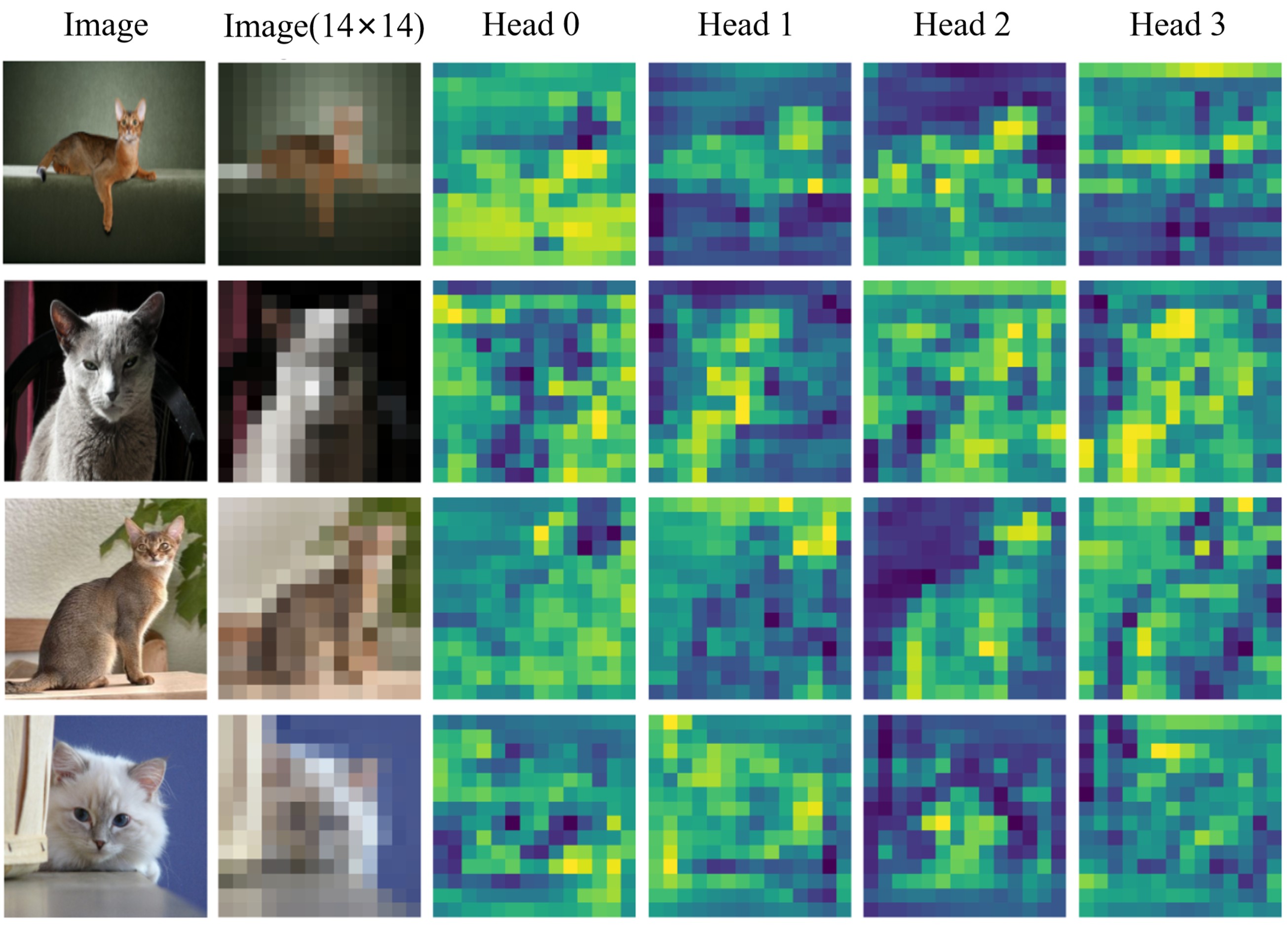}}
    \caption{Estimated memberships $\boldsymbol{\Pi}$ from DMST-T, estimated in layer 11 of each input image.}
    \label{fig:pi}
\end{center}
\end{figure}

To analyze the \(\boldsymbol{\Pi}\) matrix, we visualize the membership matrix \(\boldsymbol{\Pi}\) from a selected layer of the model. Theoretically, \(\boldsymbol{\Pi}\) enables soft clustering of $N$ tokens into $K$ subspaces through $K$ attention heads, where the entry at position \((n, k)\) denotes the estimated probability of assigning the $n$-th token to the $k$-th subspace. For visualization, we extract \(\boldsymbol{\Pi}\) from the final layer of the DMSA-tiny model, reshape it to the original image resolution, and generate heatmap visualizations. As shown in Figure \ref{fig:pi}, foreground image patches are distinctly clustered within specific subspaces, which means the learned \(\boldsymbol{\Pi}\) matrix effectively groups tokens by their subspace membership. 

\section{Conclusion}
In this paper, we propose an interpretable sparse linear attention architecture to resolve redundant encoding caused by the subspace-membership coupling in the $\mathrm{MCR}^2$ objective. 
Specifically, we first decouple the functional relationship between the membership matrix $\boldsymbol{\Pi}$ and subspaces $\boldsymbol{U}_{[K]}$, then learn the sparse $\boldsymbol{U}_{[K]}$ from a full space $\boldsymbol{S}$ via $\boldsymbol{\Pi}$; these two modifications collectively yield an optimized $\mathrm{MCR}^2$ objective. 
Unfolding this optimized objective yields DMSA. Beyond preserving linear sparse attention, DMSA’s variants inherently connect to gated attention mechanisms, establishing a theoretical bridge between white-box optimization principles and heuristic gated attention designs. Extensive visual task experiments show that the DMST(integrating DMSA into ToST as DMST) outperforms ToST by 1.08\%–1.45\% across parameter scales, with faster \(\mathrm{MCR}^2\) convergence. Visualizations furthur confirm the learned \(\boldsymbol{\Pi}\) matrix clusters tokens into semantically meaningful subspaces. 

\bibliography{example_paper}
\bibliographystyle{IEEEtran}


\appendices
\onecolumn

\section{}
\label{appendix:Theorem1}
Here, we give a detail proof of Theorem \ref{thm:vari-mcr2}.
To begin with, we need to compare the following expression with zero:
\begin{equation}
\begin{split}
& \Delta R_{\mathrm{cf}}^{\mathrm{var}}(\boldsymbol{Z} \mid (\boldsymbol{\pi}_k, \boldsymbol{U}_k^S))-\Delta R_{\mathrm{cf}}^{\mathrm{var}}(\boldsymbol{Z}, \boldsymbol{\pi}_k \mid \boldsymbol{U}_k) \\
&= R_{\mathrm{cf}}^{\mathrm{var}}(\boldsymbol{Z}, \boldsymbol{\pi}_k \mid \boldsymbol{U}_k) - R_{\mathrm{cf}}^{\mathrm{var}}(\boldsymbol{Z} \mid (\boldsymbol{\pi}_k, \boldsymbol{U}_k^S)) \\
&= \frac{1}{2} \sum_{k=1}^K \frac{n_k}{n} \sum_{i=1}^d \Bigg[ f\left( \frac{1}{n_k} \left( \boldsymbol{U}_k^\top \mathrm{Diag}(\boldsymbol{\pi}_k) \boldsymbol{Z}^\top \boldsymbol{U}_k \right)_{ii} \right)  - f\left( \frac{1}{n_k} \left( (\boldsymbol{U}_k^S)^\top \mathrm{Diag}(\boldsymbol{\pi}_k) \boldsymbol{Z}^\top \boldsymbol{U}_k^S \right)_{ii} \right) \Bigg]
\end{split}
\end{equation}

Since \(f(x) = \log\left(1 + \frac{d}{e}x\right)\) is a monotonically increasing function, we can determine redundancy by comparing its arguments. It can be proven that:

\begin{equation}
\begin{split}
&\boldsymbol{U}_k^\top \boldsymbol{Z} \mathrm{Diag}\left(
\begin{bmatrix}
\mathrm{softmax}\left( \frac{1}{2\eta} \begin{bmatrix} \|\boldsymbol{U}_1^\top \boldsymbol{Z}\|_2^2 \\ \vdots \\ \|\boldsymbol{U}_K^\top \boldsymbol{Z}\|_2^2 \end{bmatrix} \right) \\
\vdots \\
\mathrm{softmax}\left( \frac{1}{2\eta} \begin{bmatrix} \|\boldsymbol{U}_1^\top \boldsymbol{Z}\|_2^2 \\ \vdots \\ \|\boldsymbol{U}_K^\top \boldsymbol{Z}\|_2^2 \end{bmatrix} \right)
\end{bmatrix}
\right) \boldsymbol{Z}^\top \boldsymbol{U}_k - (\boldsymbol{U}_k^S)^\top \mathrm{ST}\left(\mathrm{Diag}(\boldsymbol{\pi}_k)\right) \boldsymbol{Z}^\top \boldsymbol{U}_k^S > 0.
\end{split}
\end{equation}

Because \((\boldsymbol{U}_k^S)^\top\) is a sparse selection of \(\boldsymbol{U}_k^\top\), and \(\mathrm{Diag}(\boldsymbol{\pi}_k) > \mathrm{ST}\left(\mathrm{Diag}(\boldsymbol{\pi}_k)\right)\), Theorem \ref{thm:vari-mcr2} is thus proven. 

Thus, the optimized objective achieves a larger coding rate reduction, enabling faster compression.

\section{}
\label{appendix:pseudocode}

\begin{algorithm}[h]
\caption{DMSA Attention Layer of DMST (PyTorch Implementation)}
\label{alg:dmsa_appendix}
\begin{lstlisting}[style=pythonappendix]
class AttentionDMSA(nn.Module):
    def __init__(self, dim, num_heads=8, qkv_bias=False):
        super().__init__()
        self.heads = num_heads
        self.qkv = nn.Linear(dim, dim, bias=qkv_bias)
        freqs_cis = precompute_freqs_cis(dim, 4096)
        self.register_buffer("freqs_cis", freqs_cis)

        self.PI_linear = nn.Linear(dim, num_heads,bias=None)  
        self.ST = SoftThresholdingOperation(dim=1, topk=4)
        self.sig=nn.Sigmoid()

        self.to_out = nn.Sequential(
            nn.Linear(dim, dim)
        )

    def forward(self, x):
        B, N, hd = x.shape
        S = rearrange(self.qkv(x), 'b n (h d) -> b h n d', h=self.heads)
        x = apply_rotary_emb(x, self.freqs_cis[:N])  
     
        PI = self.PI_linear(x) # Project input to membership matrix (B, N, H)
        gate = PI.mean(dim=1) # (B, H) 
        head_mask=self.ST(gate)
        
        head_mask= rearrange(head_mask, 'b h -> b h 1 1')  
        w = S * head_mask # Sparse weighting of full space S via head mask

        Pi = self.sig(rearrange(self.PI_linear(x), 'b n h -> b h n'))  
        dots = torch.matmul((Pi / (Pi.sum(dim=-1, keepdim=True) + 1e-8)).unsqueeze(-2), w ** 2) 
        attn = 1. / (1 + dots)
        out = - torch.mul(w.mul(Pi.unsqueeze(-1)), attn)
        out = rearrange(out, 'b h n d -> b n (h d)')

        return self.to_out(out)
\end{lstlisting}
\end{algorithm}

The pseudocode of the DMSA in alforithm \ref{alg:dmsa_appendix}.

\section{}
\label{appendix:training sets}
\textbf{Pre-training on ImageNet-1k}. We train our models using the AdamW optimizer with a learning rate of $2\times10^{-4}$ for 400 epochs throughout our pre-training experiments. We configure our batch size to be 512 for all our training experiments. All images are reshaped into resolutions of 224 × 224 and we tokenize each patch of size 16 × 16 into patch embeddings. For the other hyperparameters including data augmentation strategies, we adopt the exact same recipe as in \cite{ToST}. Please refer to this prior work for details. We conduct all pre-training experiments on 8 TITAN RTX 24 GB GPUs.

\textbf{Fine-tuning.} We conduct transfer learning experiments by using the pretrained TOST and DMST models as initialization and fine-tuning them on the following target datasets: CIFAR10/100, Oxford-IIIT-Pets, Food-101 and SVHN. We use batch size of 256 throughout our fine-tuning experiments. For other settings, we using the AdamW optimizer with a learning rate of $1\times10^{-3}$ for 200 epochs, weight decay = $5\times10^{-2}$. Regarding computational resources, we conducted experiments on four NVIDIA RTX 3090 GPUs

\end{document}